\documentclass[sigconf]{acmart}
\AtBeginDocument{%
  \providecommand\BibTeX{{%
    \normalfont B\kern-0.5em{\scshape i\kern-0.25em b}\kern-0.8em\TeX}}}

\setcopyright{acmcopyright}
\copyrightyear{2022}
\acmYear{2022}
\acmDOI{XXXXXXX.XXXXXXX}

\acmConference[DSHealth '22]{2022 KDD Workshop on Applied Data Science for Healthcare}{August 14--17,
  2022}{Washington DC}
\acmBooktitle{2022 KDD Workshop on Applied Data Science for Healthcare} 
\acmPrice{15.00}
\acmISBN{978-1-4503-XXXX-X/18/06}

\usepackage{lipsum}
\usepackage{multirow}
\usepackage{booktabs,tabularx,caption}
\usepackage{float}
\usepackage{amsmath,amsfonts}
\usepackage[font=footnotesize,labelfont=bf]{caption}
\usepackage{bm}

\begin{document}

%%
%% The "title" command has an optional parameter,
%% allowing the author to define a "short title" to be used in page headers.
\title{Boosting the interpretability of clinical risk scores with intervention predictions}

%%
%% The "author" command and its associated commands are used to define
%% the authors and their affiliations.
%% Of note is the shared affiliation of the first two authors, and the
%% "authornote" and "authornotemark" commands
%% used to denote shared contribution to the research.
\author{Eric Loreaux}
\email{eloreaux@google.com}
%\orcid{1234-5678-9012}
\authornotemark[1]
\affiliation{%
  \institution{Google Research}
  %\streetaddress{P.O. Box 1212}
  \city{Palo Alto}
  \state{CA}
  \country{USA}
  %\postcode{43017-6221}
}

\author{Ke Yu}
\authornote{Both authors contributed equally to this research.}
\authornote{Work completed during an internship at Google.}
\email{yu.ke@pitt.edu}
\orcid{0000000198825729}
\affiliation{%
  \institution{University of Pittsburgh}
  %\streetaddress{P.O. Box 1212}
  \city{Pittsburgh}
  \state{PA}
  \country{USA}
  %\postcode{43017-6221}
}

\author{Jonas Kemp}
\email{jonasbkemp@google.com}
%\orcid{1234-5678-9012}
\affiliation{%
  \institution{Google Research}
  %\streetaddress{P.O. Box 1212}
  \city{Palo Alto}
  \state{CA}
  \country{USA}
  %\postcode{43017-6221}
}

\author{Martin Seneviratne}
% \email{martsen@google.com}
%\orcid{1234-5678-9012}
\affiliation{%
  \institution{Google Research}
  %\streetaddress{P.O. Box 1212}
  \city{London}
  %\state{CA}
  \country{UK}
  %\postcode{43017-6221}
}

\author{Christina Chen}
% \email{christinium@google.com}
%\orcid{1234-5678-9012}
\affiliation{%
  \institution{Google Research}
  %\streetaddress{P.O. Box 1212}
  \city{Palo Alto}
  \state{CA}
  \country{USA}
  %\postcode{43017-6221}
}

\author{Subhrajit Roy}
% \email{subhrajitroy@google.com}
%\orcid{1234-5678-9012}
\affiliation{%
  \institution{Google Research}
  %\streetaddress{P.O. Box 1212}
  \city{London}
%   \state{}
  \country{United Kingdom}
  %\postcode{43017-6221}
}

\author{Ivan Protsyuk}
% \email{iprotsyuk@google.com}
%\orcid{1234-5678-9012}
\affiliation{%
  \institution{Google Research}
  %\streetaddress{P.O. Box 1212}
  \city{London}
%   \state{}
  \country{United Kingdom}
  %\postcode{43017-6221}
}

\author{Natalie Harris}
% \email{natalieharris@google.com}
%\orcid{1234-5678-9012}
\affiliation{%
  \institution{Google Research}
  %\streetaddress{P.O. Box 1212}
  \city{London}
%   \state{}
  \country{United Kingdom}
  %\postcode{43017-6221}
}

\author{Alexander D'Amour}
% \email{alexdamour@google.com}
%\orcid{1234-5678-9012}
\affiliation{%
  \institution{Google Research}
  %\streetaddress{P.O. Box 1212}
  \city{Cambridge}
  \state{MA}
  \country{USA}
  %\postcode{43017-6221}
}

\author{Steve Yadlowsky}
% \email{yadlowsky@google.com}
%\orcid{1234-5678-9012}
\affiliation{%
  \institution{Google Research}
  %\streetaddress{P.O. Box 1212}
  \city{Cambridge}
  \state{MA}
  \country{USA}
  %\postcode{43017-6221}
}

\author{Ming-Jun Chen}
% \email{mingjunchen@google.com}
%\orcid{1234-5678-9012}
\affiliation{%
  \institution{Google Research}
  %\streetaddress{P.O. Box 1212}
  \city{Palo Alto}
  \state{CA}
  \country{USA}
  %\postcode{43017-6221}
}

%%
%% By default, the full list of authors will be used in the page
%% headers. Often, this list is too long, and will overlap
%% other information printed in the page headers. This command allows
%% the author to define a more concise list
%% of authors' names for this purpose.
\renewcommand{\shortauthors}{Loreaux and Yu, et al.}

%%
%% The abstract is a short summary of the work to be presented in the
%% article.
\begin{abstract}
Machine learning systems show significant promise for forecasting patient adverse events via risk scores. However, these risk scores implicitly encode assumptions about future interventions that the patient is likely to receive, based on the intervention policy present in the training data. Without this important context, predictions from such systems are less interpretable for clinicians. We propose a joint model of intervention policy and adverse event risk as a means to explicitly communicate the model's assumptions about future interventions. We develop such an intervention policy model on MIMIC-III, a real world de-identified ICU dataset, and discuss some use cases that highlight the utility of this approach. We show how combining typical risk scores, such as the likelihood of mortality, with future intervention probability scores leads to more interpretable clinical predictions.
\end{abstract}

%%
%% The code below is generated by the tool at http://dl.acm.org/ccs.cfm.
%% Please copy and paste the code instead of the example below.
%%


%%
%% Keywords. The author(s) should pick words that accurately describe
%% the work being presented. Separate the keywords with commas.

%% A "teaser" image appears between the author and affiliation
%% information and the body of the document, and typically spans the
%% page.
%%\begin{teaserfigure}
%%  \includegraphics[width=\textwidth]{sampleteaser}
%%  \caption{Seattle Mariners at Spring Training, 2010.}
%%  \Description{Enjoying the baseball game from the third-base
%%  seats. Ichiro Suzuki preparing to bat.}
%%  \label{fig:teaser}
%%\end{teaserfigure}

%%
%% This command processes the author and affiliation and title
%% information and builds the first part of the formatted document.
\maketitle

\section{Introduction}
The digitization of medical data has created many opportunities for machine learning systems to empower clinicians and improve medical care \citep{ghassemi2020challengesopportunities}. One key area of interest is the use of predictive models to quantify risk of mortality or deterioration, broadly referred to as acuity. Most established patient acuity scores  depend on a small number of physiological risk factors determined through consensus of various medical bodies, e.g. APACHE~\citep{apacheIII1991}, SOFA~\citep{vincent1996sofa}, and SAPS II~\citep{le1993new}. In recent years, deep learning-based frameworks have improved upon the accuracy of these scores by replacing simple risk factors with a wider breadth of patterns that are not as easily detectable by human clinicians~\citep{shickel2017deep}.

\begin{figure}[t]
    \centering
    \includegraphics[width=0.49\textwidth]{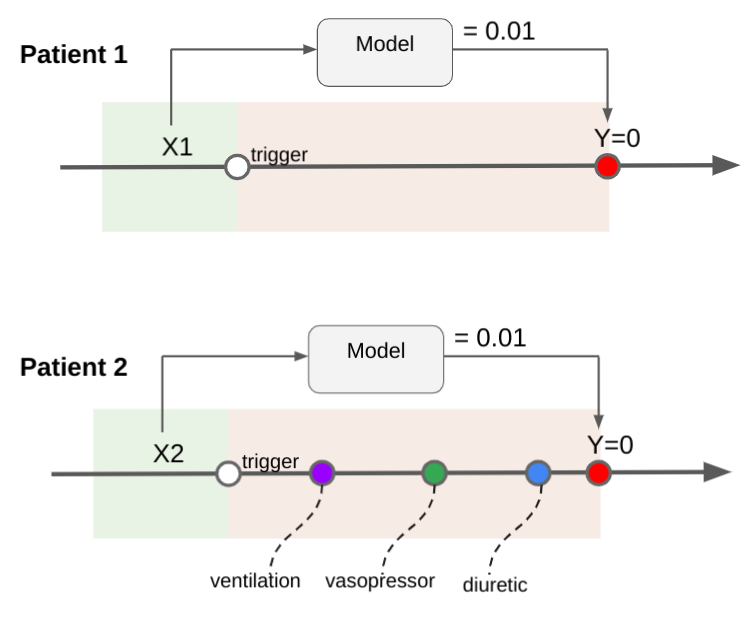}
    
    \caption{Two patients whose predicted risks are equally low ($1\%$), but whose clinical courses differ. For patient 2, the prediction may be predicated on assumptions about future treatments. We aim to communicate these assumptions.}
    
    \label{fig:motivation}
\end{figure}

While more powerful risk scores are promising, critical deployment challenges  remain, including interpretability ~\citep{kelly2019challenges}. Most clinical interpretability methods are aimed at explaining how input features influence model predictions, but these explanations fail to communicate an important piece of clinical context: the model's implicit, learned assumptions about likely future interventions given a patient's current state. Fig.~\ref{fig:motivation} visualizes the problem of two equivalently low risk predictions being predicated upon completely different presumed intervention trajectories, which cannot be communicated in a standalone acuity score. Such implicit assumptions have been shown to negatively impact model behavior in real-world deployment scenarios. For example, \cite{caruana2015pneumoniarisk} discusses a case in which a model trained to produce mortality risk scores for pneumonia patients learns to ascribe lower risk to patients with asthma, when in fact the opposite should be true. This behavior is explained by the fact that asthma patients receive more aggressive and effective interventions than other patients. This case highlights the need for interpretability beyond simple feature attribution, in which the assumptions about future interventions are made clearer. 

In this work, we demonstrate that communicating intervention assumptions in addition to the patient’s future risk has the potential to improve interpretability and increase trust among end users considering said risk scores in care planning. In order to demonstrate the utility of this risk score contextualization, we develop a recurrent multitask neural network which simultaneously predicts the likelihood of both mortality and a collection of relevant clinical interventions. We evaluate this approach on the MIMIC-III dataset and highlight examples in which the additional context of future intervention probabilities can enable a machine learning system to produce more interpretable predictions for clinicians.

\section{Related Work}
It is well known that patient risk is highly dependent on intervention. \cite{lenert2019progmodsvictimsusccess} simulated adverse event label drift in response to changing intervention, and recommend incorporating intervention predictions as a solution, which we explore further in this work. \cite{vangeloven2020progmodsvictimsusccess} formally defined a unified endpoint combining adverse events and interventions after baseline. Rather than change the definition of the risk being predicted, we seek to use predictions of future interventions to better contextualize this risk. 

We are by no means the first to propose the prediction of multiple clinical interventions. \cite{ghassemi2017switchstatespaceinterventions} employed a switching state space model to represent patient states and evaluated the performance of learned states in prediction of five ICU treatments. \cite{suresh2017clinical} integrated multiple EHR data sources and used the learned representation to predict onset and weaning of five interventions. \cite{wang2020mimic} built a pipeline to extract continuous intervention signals for ventilators, vassopressors, and fluid therapies. We extend upon previous intervention prediction tasks (i.e., to thirteen interventions) and specifically demonstrate how such predictions can be used to increase the interpretability of patient acuity scores.

There have been many efforts to improve the interpretability of clinical machine learning systems. \cite{payrovnaziri2020interpreview} provide a comprehensive overview of previous work. Most of these techniques focus on attributing model outputs to individual features, or to clinical concepts (e.g. \cite{mincu2021tcavehr}). We diverge from this work by communicating the model's implicit assumptions about future events that may have a causal relationship with the predicted endpoint.

\section{Data and Label Description}
\label{sec:data}
\textbf{Data} We used MIMIC-III~\citep{johnson2016mimic} for our experiments. MIMIC-III is a publicly available dataset that includes the de-identified electronic health records (EHRs) of 53,423 patients admitted to critical care units at the Beth Israel Deaconess Medical Center from 2001 to 2012.  We selected the study cohort based on three inclusion criteria: (1) the patient was at least 15 years old at the time of admission; (2) following \cite{wang2020mimic}, we only considered each patient's first ICU stay to prevent possible information leakage; (3) the total duration of the ICU stay was between 2 and 14 days. The final cohort included 18,335 ICU stays and was randomly split into training (80\%), validation (10\%), and test (10\%). Table~\ref{tbl:demo} describes the cohort demographics. 

\noindent\textbf{Label description} We focused on the early prediction of interventions as auxiliary tasks, jointly trained with in-ICU mortality prediction. All predictions are binary classification tasks that use all historical information up to and including the first 24 hours of a patient's first ICU stay, and a prediction horizon that extends to the end of that same ICU stay. We targeted 13 commonly used interventions in ICU, including ICU medications (e.g., vasopressors), fluid boluses (e.g., crystalloid bolus), transfusions (e.g., RBC transfusion), and mechanical ventilation. To the best of our knowledge, no other work has included as many simultaneous intervention prediction tasks.  Definitions of interventions were based on the SQL queries available in the MIMIC-III codebase~\citep{johnson2018mimic} and the item IDs of corresponding concepts were reviewed and modified with clinical guidance where appropriate (see Appdx.~\ref{apdx:intv_def} for detailed definitions).  The in-ICU mortality label was derived using the ``DEATHTIME'' included in the MIMIC-III ``ADMISSION'' table. A half-hour tolerance period was added to the ``OUTTIME'' in the ``ICUSTAYS'' table to account for patients who died within the 30 minutes of ICU discharge. Table~\ref{tbl:label} shows the prevalence of in-ICU mortality as well as all the intervention labels split by train, validation, and test data sets.

\section{Methodology}
\subsection{Baseline Models}
%The initial intent of risk score models is to forecast ICU mortality; hence, they may not be suitable for intervention prediction. Deep learning models, on the other hand, are capable of automatically learning task-relevant features. In order to ensure that a deep recurrent neural network provides additional value for the proposed prediction tasks, 
We built two logistic regression models based on features used in SOFA and SAPS-II, respectively. The SOFA baseline model included \textit{six} predictors that measure the rates of failure of six organ systems. The SAPS-II baseline model included \textit{seventeen} variables, including routine physiological measurements, comorbidity, and admission status. Both SOFA and SAPS-II scores are commonly used to estimate the probability of mortality for ICU patients. In this study, we investigate their effectiveness in proposed prediction tasks, including both mortality and intervention predictions. 

\subsection{LSTM-based multi-label learning model}
We extended an open-source implementation\footnote{https://github.com/google/ehr-predictions} of the LSTM-based mortality predictive model from \cite{tomavsev2021use} to incorporate multi-label intervention prediction tasks as defined in Sec.~\ref{sec:data}. The model used a version of MIMIC-III mapped to the Fast Healthcare Interoperability Resource (FHIR) standard as described in~\cite{rajkomar2018scalable}. Every patient's record was represented as a collection of timestamped events with an associated clinical code and, where applicable, a value. We obtained the feature representation from FHIR resources as described in \cite{50294}. %The main steps are: (1) all numerical features were normalized to the [0, 1] range after excluding extreme values (lower than 1st or higher than 99th percentile); (2) features were aggregated into hourly buckets, and median is used for repeated values; (3) presence of the continuous features were explicitly encoded to distinguish between the absence of a numerical value and an actual value of zero; (4) categorical features were represented as one-hot vectors. The full feature set consists of \todo{32,170} continuous variables and \todo{38,600} categorical variables. 

\noindent \textbf{Loss Function} We employed a multi-label learning approach to train the model $f$. A patient's input tensor $x^{(i)}$ is fed through a sparse embedding layer followed by an LSTM~\citep{hochreiter1997long} module parameterized by $\theta$. The output vector of the LSTM (i.e., hidden patient representation) is then shared across fourteen different classification heads, each of which is parameterized by $\varphi_j$. We used binary cross-entropy as the loss function for each classification task and minimize their sum to train the entire network. The complete loss function is:
%\vspace{-5px}
\begin{align*}
    \mathcal{L} = - & \sum_j^P \sum_i^N y_j^{(i)}\log f(x^{(i)};\theta, \varphi_j) \\
   & + (1-y_j^{(i)})  \log \big(1- f(x^{(i)};\theta, \varphi_j)\big),
\end{align*}
where $N$ is the number of training samples and $P$ is the total number of classification tasks. In our experiments, $P=14$ which includes \textit{one} in-ICU mortality prediction task and \textit{thirteen} intervention prediction tasks. For implementation details, we refer the reader to Appdx.~\ref{apdx:impl}.

\begin{figure*}[t!]
    \centering
    \includegraphics[width=0.95\textwidth]{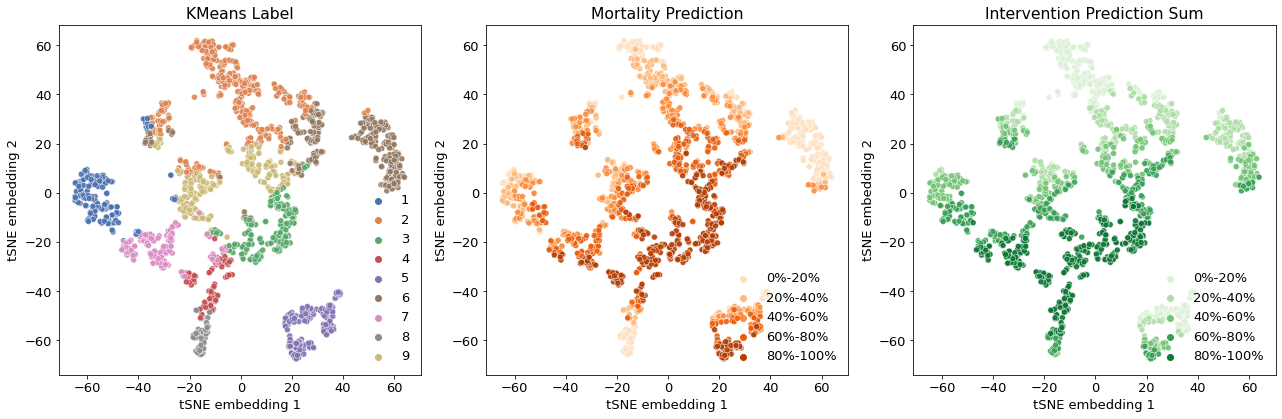}
    
    \caption{2D t-SNE visualization of the LSTM-based intervention predictions. Each point represents one patient. The left panel is labeled by K-means cluster assignment ($K=9$) using the 13 intervention predictions as inputs. The middle panel is labeled by the quantile group of predicted mortality risk. The right panel is labeled by the quantile group of total predicted intervention scores, summed across the 13 interventions.}
    
    \label{fig:t-SNE}
\end{figure*}

\section{Experiments and Results}

We first present quantitative results, then show both aggregate and individual use cases that demonstrate the benefits of pairing mortality prediction with intervention predictions. All experimental results are based on the held out test set.
%(1) a blind spot in the standalone mortality prediction model with intervention predictions. 

\subsection{Model Results and Comparison}
\begin{table}[t]
\caption{Model performance comparison. Bold face indicates the highest average AUC scores.}
\scalebox{0.7}{
\begin{tabularx}{0.7\textwidth}{@{\extracolsep{\fill}}lcccccc@{\extracolsep{\fill}}}
\toprule \addlinespace
\multirow{2}{*}{\textbf{Task}} & & \textbf{AUROC} &&& \textbf{AUPRC}& \\ \addlinespace
\cmidrule(r){2-4} \cmidrule(l){5-7}
& SOFA & SAPS II & LSTM &  SOFA & SAPS II & LSTM \\
\addlinespace
\midrule
In-ICU Mortality & $0.69$ & $0.76$ & $\bm{0.82 \pm 0.01}$ & $0.23$ & $0.27$  & $\bm{0.39 \pm 0.03}$\\
Vasopressors & $0.66$ & $0.70$ & $\bm{0.74 \pm 0.00}$ & $0.30$ & $0.31$ & $\bm{0.37 \pm 0.01}$\\
Inotropes & $0.70$ & $0.69$ & $\bm{0.77 \pm 0.02}$ & $0.10$ & $0.10$ & $\bm{0.16 \pm 0.01}$\\
Sedation & $0.57$ & $0.65$ & $\bm{0.76 \pm 0.01}$ & $0.33$ & $0.38$ & $\bm{0.50 \pm 0.01}$\\
Analgesic & $0.57$ & $0.66$ & $\bm{0.85 \pm 0.01}$ & $0.34$ & $0.38$ & $\bm{0.64 \pm 0.02}$\\
Anticoagulation & $0.59$ & $0.58$ & $\bm{0.86 \pm 0.01}$ & $0.38$ & $0.36$ & $\bm{0.77 \pm 0.02}$ \\
Diuretic & $0.65$ & $0.68$ & $\bm{0.88 \pm 0.01}$ & $0.29$ & $0.30$ & $\bm{0.68 \pm 0.02}$\\
Paralytic & $0.71$ & $0.86$ & $\bm{0.96 \pm 0.02}$ & $0.06$ & $0.18$ & $\bm{0.27 \pm 0.13}$\\
Colloid Bolus & $0.66$ & $0.72$ & $\bm{0.78 \pm 0.01}$ & $0.12$ & $0.18$ & $\bm{0.26 \pm 0.02}$\\
Crystalloid Bolus & $0.57$ & $0.61$ & $\bm{0.65 \pm 0.01}$ & $0.48$ & $0.51$ & $\bm{0.55 \pm 0.01}$\\
FFP Transfusion & $\bm{0.75}$ & $0.67$ & $0.71 \pm 0.01$ & $\bm{0.24}$ & $0.16$ & $0.17 \pm 0.03$\\
RBC Transfusion & $0.65$ & $0.68$ & $\bm{0.76 \pm 0.02}$ & $0.45$ & $0.45$ & $\bm{0.57 \pm 0.02}$\\
Ventilation & $0.53$ & $0.55$ & $\bm{0.63 \pm 0.01}$ & $0.14$ & $0.15$ & $\bm{0.21 \pm 0.02}$\\
Antibiotic & $0.58$ & $0.56$  & $\bm{0.90 \pm 0.00}$ & $0.88$ & $0.88$ & $\bm{0.98 \pm 0.00}$\\
\bottomrule
\end{tabularx}
}
\centering
\label{tab:auc}
\end{table}

We evaluated the model using AUROC and AUPRC scores and present the values in Tab.~\ref{tab:auc}. We find that, compared to the baseline models using severity score features, the LSTM-based model yields improvements on most intervention predictions, except for two (paralytic, FFP transfusion) with very low label prevalence. %A mortality risk score is often used as a surrogate for resource management in ICU. 
We also find that compared to the prediction of in-ICU mortality, four intervention predictions have higher AUROC scores and eight have higher AUPRC scores, suggesting that these predictions can offer more specific and reliable signals for predicting patient needs than a mortality risk score.

\begin{table}[h]
\caption{Mean values and standard deviations for precision by the number of interventions ($I$) that patient were given. Bold face indicates the highest mean values. Precision at $I$ is the proportion of top $I$ predicted interventions that are in the patient's recorded intervention set.}
\scalebox{0.8}{
\begin{tabularx}{0.6\textwidth}{@{}XXXX@{}}
\toprule 
\addlinespace
\multirow{2}{*}{\textbf{Interventions}} & & \textbf{Precision at $I$} & \\
\cmidrule(r){2-4}
& SOFA & SAPS II & LSTM \\ 
\midrule
$I = 1$ & $0.84 \pm 0.37$ & $ 0.84 \pm 0.37$ & $ \textbf{0.89} \pm \textbf{0.31}$ \\
$I = 2$ & $0.63 \pm 0.26$ & $0.65 \pm 0.25$ & $\textbf{0.76} \pm \textbf{0.26}$ \\
$I = 3$ & $0.62 \pm 0.22$ & $0.64 \pm 0.21$ & $\textbf{0.73} \pm \textbf{0.20}$ \\
$I = 4$ & $0.61 \pm 0.18$ & $0.62 \pm 0.18$ & $\textbf{0.71} \pm \textbf{0.18}$ \\
$I = 5$ & $0.66 \pm 0.18$ & $0.68 \pm 0.18$ & $\textbf{0.76} \pm \textbf{0.15}$ \\
$I \geq 6$ & $0.77 \pm 0.13$ & $0.77 \pm 0.13$ & $\textbf{0.81} \pm \textbf{0.12}$ \\
\bottomrule
\end{tabularx}
}
\centering
\label{tab:precision}
\end{table}

We evaluated the model's capability of ranking the most relevant interventions for each patient. We grouped patients by the number of interventions (denoted as $I$) that they received and evaluated our model with Precision\textit{@}$I$ metrics, which is the proportion of top $I$ predicted interventions that are in the patient's recorded intervention set. Tab.~\ref{tab:precision} shows that our LSTM-based model yields fairly high retrieval precision, with mean values $>0.7$, at all levels, and compares favorably to baseline models.

We also evaluated the models for calibration performance and for the relationship between in-ICU mortality prediction and intervention predictions. We present these results in Appdx.~\ref{sec:supexp}.

\begin{figure*}[t]
    \centering
    \includegraphics[width=0.8\textwidth]{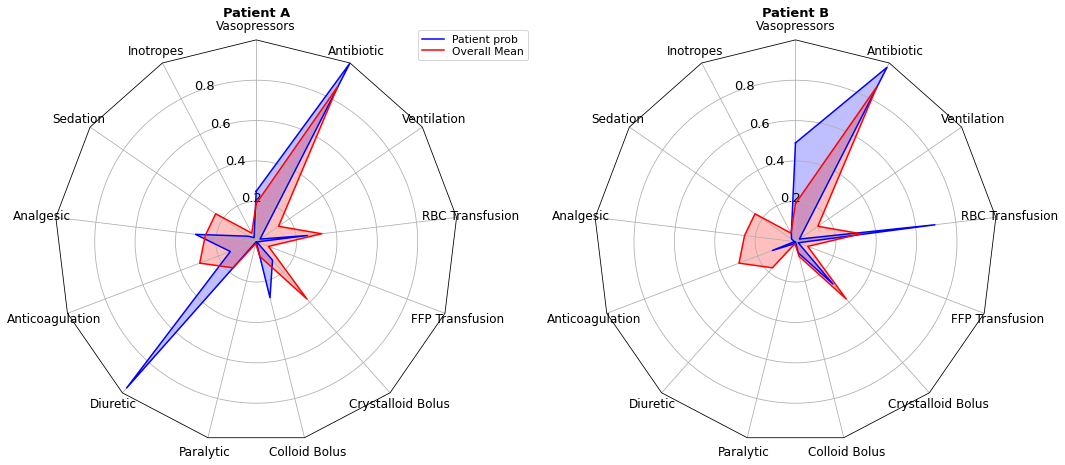}
    
    \caption{Comparison of the future intervention probabilities of patients A and B. Both have similar medical profiles and predicted mortality risk, and yet their very different predicted interventions indicate that their mortality risk is predicated on distinct assumed clinical trajectories.
}
    
    \label{fig:patcomp}
\end{figure*}

\subsection{Use Cases}
\noindent \textbf{Cluster analysis:} We created a t-SNE~\citep{laurens2008visualizing} plot using the thirteen intervention predictions of the LSTM-based model as inputs to obtain the 2D t-SNE embedding. Fig.~\ref{fig:t-SNE} shows the same t-SNE plot labeled by three different categorical variables: K-means cluster assignment using the entire future intervention probability vector, quantile group of predicted mortality risk, and quantile group of total predicted intervention score. 

Cluster 8 (left panel, grey points at bottom) in Fig.~\ref{fig:t-SNE} represents a set of patients who have low predicted mortality risk (center panel, light red) but require extensive medical interventions (right panel, dark green). Upon clinical review, we identified these patients as cardiothoracic surgery patients (coronary artery bypass grafts (CABG), valve replacements, etc.) receiving post-surgical care. Fig.~\ref{fig:radar} shows that patients in cluster 8 were more likely to receive diuretics to prevent fluid overload and antibiotics to manage post-surgical infection, but were less likely to be sedated or ventilated in the ICU. These predictions indicate to the clinician that the patient risk score they are viewing is already predicated on this future post-surgical trajectory, without which, risk may have been much higher.

% \noindent \textbf{Case 2:} Fig.~\ref{fig:t-SNE} shows that patients with the highest mortality risks (dark red color) are concentrated in clusters 3 and 4. However, as shown in Fig.~\ref{fig:radar}, these two clusters exhibit different intervention profiles. For example, patients in cluster 4 are more likely to receive analgesics, anticoagulation, colloid boluses, and diuretics, while patients in cluster 3 are more likely to receive fresh frozen plasma (FFP) transfusion and crystalloid boluses. While the model predicts similarly high risk for these two patient clusters, each cluster's risk is predicated on distinct treatment plans, enabling an additional interpretable stratification of patient forecasts.

\noindent \textbf{Patient trajectories:} To illustrate the potential of intervention forecasting to boost interpretability at the patient level, we selected two patients (A and B) with similar risk scores but differing intervention predictions. Both were 80-85 year old males admitted to the ICU after multi-vessel CABG; both were extubated on day 1 post-operatively with 24h risk scores in the bottom tenth percentile. Fig.~\ref{fig:patcomp} shows that despite similar risk, Patient A had a higher probability of receiving diuretics, whereas Patient B had a significantly higher probability of requiring transfusion, crystalloid bolus and vasopressors. This forecast matches the differing ICU trajectories of these patients. While patient A's admission was complicated by tachycardia and hypertension, patient B had a more complicated admission requiring a post-op transfusion, vasopressors after a hypotensive episode, chemical cardioversion and anticoagulation for new atrial fibrillation, and intravenous antibiotics for phlebitis. These additional signals indicate to the clinician that the very similar risk scores for these two similar patients are actually predicated on distinct clinical trajectories.

\section{Conclusion and Future Work}
In this work, we demonstrate ways in which modeling intervention policy can improve the interpretability of clinical risk scores. Providing this context has the potential to increase confidence and trust as well as improve inter-provider variability in practice through personalized patient forecasts. This interpretability technique is applicable to any prediction task in which the predictions depend on future actions which may have a causal relationship with the target endpoint. It is especially critical to communicate these assumptions when the consumer of the predictions is also implicated in these future actions. 

In future work, we hope to expand upon the findings presented here by applying our policy model to health environments beyond the ICU, such as in general inpatient hospital wards or in the outpatient setting. The heterogeneity of care trajectories is wider in these settings, and so the potential ambiguity regarding patient risk forecasts is amplified. In addition to expanding to new datasets, we would like to explore a number of improvements to the quality and granularity of our intervention policy models, such as shifting to continuous predictions and increasing the number and sophistication of intervention patterns forecasted. It will also be imperative to conduct user testing with clinicians to determine clinical benefit and potential for integration into real-world workflows.

%%
%% The acknowledgments section is defined using the "acks" environment
%% (and NOT an unnumbered section). This ensures the proper
%% identification of the section in the article metadata, and the
%% consistent spelling of the heading.
%\begin{acks}
%To Robert, for the bagels and explaining CMYK and color spaces.
%\end{acks}

%%
%% The next two lines define the bibliography style to be used, and
%% the bibliography file.
\bibliographystyle{ACM-Reference-Format}
\bibliography{kdd-main}

%%
%% If your work has an appendix, this is the place to put it.
\appendix
\onecolumn
\renewcommand\thefigure{\thesection.\arabic{figure}}    
\renewcommand\thetable{\thesection.\arabic{table}}    
\section{Intervention Label Definition}
\label{apdx:intv_def}
\begin{itemize}
    \item \textbf{Vasopressors}: Vasopressors are a group of medications used to treat severely low blood pressures by constricting blood vessels. The vasopressors label was defined as the onset of any of the following 7 drugs: levophed, neosynephrine, phenylephrine, norepinephrine, vasopressin, dopamine, or epinephrine.
    \item \textbf{Inotropes}: Inotropes are medications that change the force of patients' hearts' contractions. In ICU, inotropes are used to stabilise patients' circulation and optimise oxygen supply. The inotropes label was defined as the onset of any of the following 3 drugs: dopamine, dobutamine, or milrinone.
    \item \textbf{Sedation}: Sedative drugs are primarily used for the treatment of agitation and anxiety caused by many different conditions, such as dyspnea, mechanical ventilation, and untreated pain. The sedation label was defined as the onset of any of the following 7 drugs: propofol, midazolam, ativan, dexmedetomidine, diazepam, ketamine or pentobarbitol.
    \item \textbf{Analgesic}: Analgesics are used for pain control. The analgesic label was defined as the onset of any of the following 3 drugs: fentanyl, morphine sulfate, or hydromorphone.
    \item \textbf{Anticoagulation}: Anticoagulation is used for the management of  venous thromboembolism (VTE), atrial fibrillation (AF), mechanical heart valves, and idiopathic pulmonary arterial hypertension (IPAH). The anticoagulation label was defined as the onset of any of the following 7 drugs: heparin, integrelin, argatroban, lepirudin, aggrastat, reopro, or bivalirudin. 
    \item \textbf{Diuretic}: Diuretics are a mainstay of treatment for managing fluid overload in ICU. The diuretic label was defined as the onset of either of the following 2 drugs: furosemide or natrecor.
    \item \textbf{Paralytic}: Paralytics, or neuromuscular blocking agents (NMBAS), paralyze skeletal muscles by blocking the transmission of nerve impulses at the myoneural junction. They are often deployed in the sickest patients in ICU when usual care fails%~\cite{price2012fresh}
    . The paralytic label was defined as the onset of any of the following 3 drugs: cisatracurium, vecuronium, or atracurium.
    \item \textbf{Colloid Bolus}: Colloids are gelatinous solutions that maintain a high osmotic pressure in the blood. Colloid boluses are used to improve cardiovascular function and organ perfusion and are often considered as less aggressive alternatives to vasopressors%~\cite{malbrain2014fluid}
    . The colloid bolus label was derived from the presence of item IDs defined in the SQL query \texttt{colloid\_bolus.sql} provided on the MIMIC Github repository\footnote{https://github.com/MIT-LCP/mimic-code/tree/main/mimic-iii}.
    \item \textbf{Crystalloid Bolus}: Crystalloids, another type of fluid bolus, are defined as solutions of ions that are capable of passing through semipermeable membranes%~\cite{finfer2018intravenous}
    . The crystalloid bolus label was derived from the presence of item IDs defined in the SQL query \texttt{crystalloid\_bolus.sql} provided on the MIMIC Github repository.
    \item \textbf{FFP Transfusion}: Fresh frozen plasma (FFP) transfusions are widely used in ICU patients to correct deficiency of coagulation factors or increased risk of developing acute lung injury%~\cite{murad2010effect
    . The FFP transfusion label was derived from the presence of item IDs defined in the SQL query \texttt{ffp\_transfusion.sql} provided on the MIMIC Github repository.
    \item \textbf{RBC Transfusion}: Red blood cell (RBC) transfusions are commonly used in ICU patients with increased mortality risk%~\cite{marik2008efficacy}
    . The RBC transfusion label was derived from the presence of item IDs defined in the SQL query \texttt{rbc\_transfusion.sql} provided on the MIMIC Github repository.
    \item \textbf{Ventilation}:  Mechanical ventilation is commonly used when an ICU patient requires assistance for breathing. The ventilation label was based on the SQL query \texttt{ventilation\_durations.sql} provided on the MIMIC Github repository, and was defined as the onset of a new mechanical ventilation event.
    \item \textbf{Antibiotic}: The antibiotic label was based on the SQL query \texttt{antibiotic.sql} provided on the MIMIC Github repository, with the following modifications: (1) we only included drugs prescribed to be given through an intravenous (IV) line, and (2) we ensured that there was a input event or chart event associated with vancomycin if there was a prescription order of vancomycin.
\end{itemize}

\newpage
\section{Implementation Details}
\label{apdx:impl}
We report the optimal hyperparameter configuration on the validation data set. We trained a recurrent neural network with 3 layers of LSTM cells of size 200 with batch size 128. For the feature embedding, we used a sparse lookup table embedding with an embedding dimension of 300. We used Xavier initialization~\citep{glorot2010understanding}, Adam optimization~\citep{kingma2014adam}, and learning rate 0.0001 with exponential decay 0.85 per 12,000 steps. We trained for a total of 200,000 steps on an NVIDIA Tesla V100 GPU. For regularization, we implemented L1 regularization on the sparse lookup embeddings with a strength of 0.0005, and also input, output, and variational recurrent dropout~\citep{gal2016varrecdropout} on all of our LSTM cells with a dropout probability of 0.4.

We also explored the use of two additional recurrent cell architectures: the gated recurrent unit (GRU)~\citep{cho2014gru} and the simple recurrent unit (SRU)~\citep{lei2017gru}, as well as two additional sparse lookup embedding sizes (200, 400) and one additional initial learning rate (0.001). While we found the LSTM to be the best performing architecture, we saw very little difference across the two learning rates and three embedding dimensions.

\newpage
\section{Data Summary Statistics}
\label{apdx:supp}

% demographic statistics 
\begin{table}[ht]
\caption{Cohort summary by demographic and admission variables.}
\begin{tabularx}{0.8\textwidth}{@{}lXXXl@{}}
\toprule \addlinespace
& & \textbf{Gender} & & \textbf{Total} \\
& & F & M & \\
\midrule
\textbf{Ethnicity} & Asian & $181$ & $258$ & $439$ $(2\%)$ \\
& Hispanic & $227$ & $329$ & $556$ $(3\%)$ \\
& Black & $773$ & $609$ & $1,382$ $(8\%)$ \\
& Other & $1,252$ & $1,751$ & $3,003$ $(16\%)$ \\
& White & $5,718$ & $7,237$ & $12,955$ $(70\%)$ \\
\midrule
\textbf{Age} & $<30$ & $366$ & $526$ & $892$ $(5\%)$ \\
& 31-50 & $1,195$ & $1,657$ & $2,852$ $(16\%)$ \\
& 51-70 & $2,699$ & $4,169$	& $6,868$ $(37\%)$ \\
& $>70$ & $3,891$ & $3,832$ &	$7,723$ $(42\%)$ \\
\midrule
\textbf{Insurance Type} & Self Pay & $54$ & $151$ & $205$ $(1\%)$ \\
& Government & $184$ & $331$ & $515$ $(3\%)$ \\
& Medicaid &	$625$ & $800$ & $1,425$ $(8\%)$ \\
& Private & $2,257$ & $3,606$ & $5,863$ $(32\%)$ \\
& Medicare & $5,031$ & $5,296$ & $10,327$ $(56\%)$ \\
\midrule
\textbf{Admission Type} & Urgent & $255$ & $291$ & $546$ $(3\%)$ \\
& Elective & $1,143$ & $1,599$ & $2,742$ $(15\%)$ \\
& Emergency	& $6,753$ & $8,294$ & $15,047$ $(82\%)$ \\
\midrule
\textbf{First Careunit} & TSICU	& $931$ & $1,435$ & $2,366$ $(13\%)$ \\
& CCU & $1,226$ & $1,603$ & $2,829$ $(15\%)$ \\
& SICU & $1,481$ & $1,540$ & $3,021$ $(16\%)$ \\
& CSRU & $1,416$ & $2,403$ & $3,819$ $(21\%)$ \\
& MICU & $3,097$ & $3,203$ & $6,300$ $(34\%)$ \\
\midrule
\textbf{Total} & & $8,151$ & $10,184$ & $18,335$ $(100\%)$ \\
\bottomrule
\end{tabularx}
\centering
\label{tbl:demo}
\end{table}

% Label prevalence summary table
\begin{table}[htbp]
\caption{Summary statistics for the labels.}
\begin{tabularx}{0.8\textwidth}{@{}XXXl@{}}
\toprule \addlinespace
& \textbf{Training} & \textbf{Validation} & \textbf{Test} \\ \addlinespace
\midrule
In-ICU Mortality & $1,329$ $(8.9\%)$ & $129$ $(7.4\%)$ & $141$ $(8.2\%)$ \\
Vasopressors & $2,486$ $(16.7\%)$ & $261$ $(15.1\%)$ & $306$ $(17.9\%)$ \\
Inotropes & $667$ $(4.5\%)$ & $70$ $(4.1\%)$ & $77$ $(4.5\%)$ \\
Sedation & $3,766$ $(24.3\%)$ & $431$ $(24.9\%)$ & $431$ $(25.2\%)$ \\
Analgesic & $3,994$ $(26.8\%)$ & $450$ $(24.9\%)$ & $436$ $(25.5\%)$ \\
Anticoagulation & $4,439$ $(29.8\%)$ & $518$ $(30.0\%)$ & $525$ $(30.7\%)$ \\
Diuretic & $2,766$ $(18.6\%)$ & $308$ $(17.8 \%)$ & $306$ $(17.9\%)$ \\
Paralytic & $130$ $(0.9\%)$ & $14$ $(0.8\%)$ & $16$ $(0.9\%)$ \\
Colloid Bolus & $928$ $(6.2\%)$ & $101$ $(5.8\%)$ & $99$ $(5.8\%)$ \\
Crystalloid Bolus & $5,897$ $(39.6\%)$ & $655$ $(37.9\%)$ & $683$ $(39.9\%)$ \\
FFP Transfusion & $783$ $(5.3\%)$ & $87$ $(5.0\%)$ & $103$ $(6.0\%)$ \\
RBC Transfusion & $4,602$ $(30.9\%)$ & $516$ $(29.9\%)$ & $524$ $(30.6\%)$ \\
Ventilation & $1,935$ $(13.0\%)$ & $212$ $(12.3\%)$ & $219$ $(12.8\%)$ \\
%Ventilation Reinitiation & 1,179 (7.9\%) & 144 (8.3\%) & 129 (7.5\%) \\
Antibiotic & $12,906$ $(86.6\%)$ & $1,503$ $(86.9\%)$ & $1,467$ $(85.7\%)$ \\
\midrule
\textbf{Total} & $14,895$ $(100\%)$ & $1,728$ $(100\%)$ & $1,712$ $(100\%)$ \\
\bottomrule
\end{tabularx}
\centering
\label{tbl:label}
\end{table}

\newpage
\section{Supplementary Experimental Results and Details}
\label{sec:supexp}

\begin{figure}[ht]
    \centering
    \includegraphics[width=1.0\textwidth]{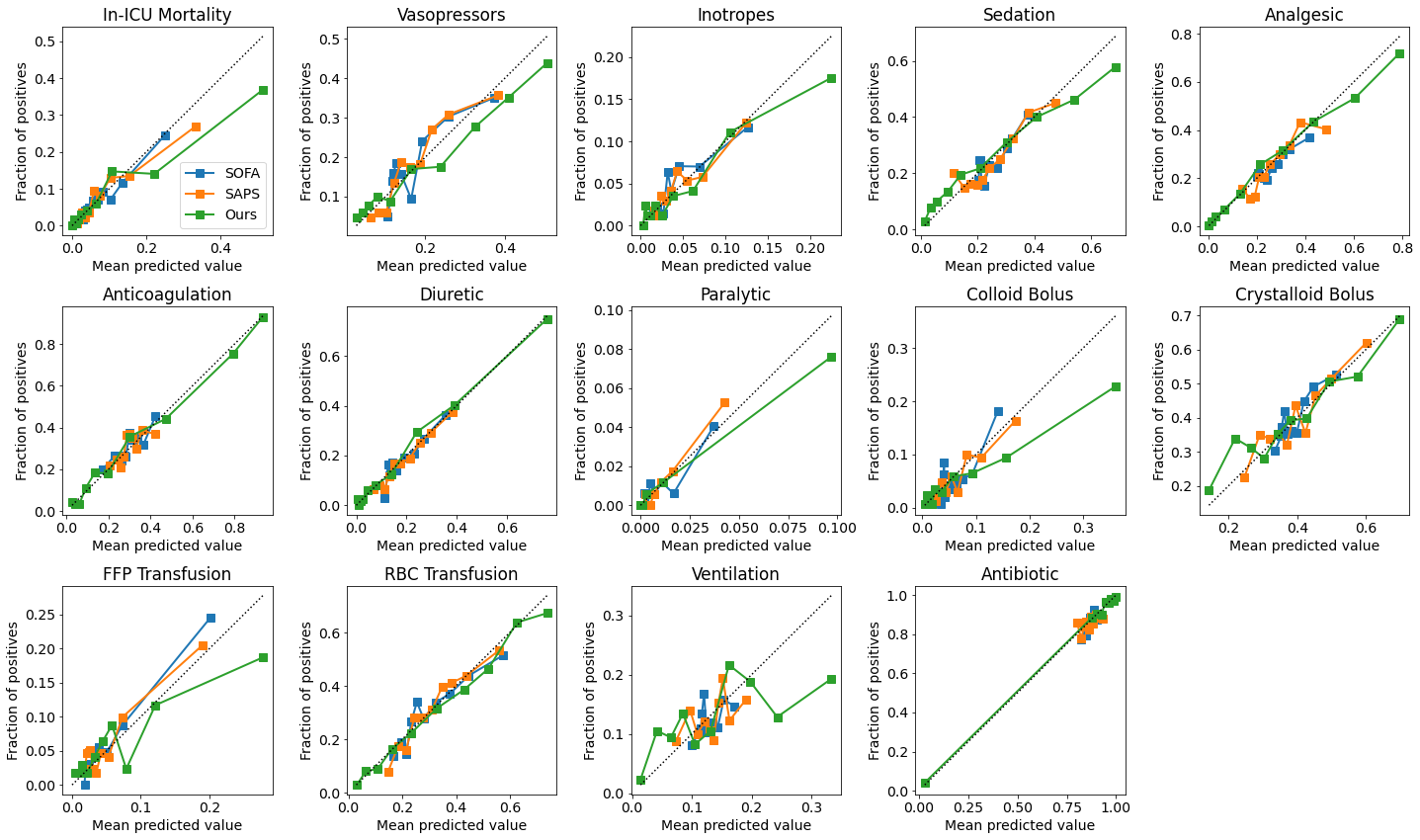}
    
    \caption{Calibration curves for in-ICU mortality and intervention predictions.}
    
    \label{fig:6}
\end{figure}

\begin{figure}[ht]
    \centering
    \includegraphics[width=1.0\textwidth]{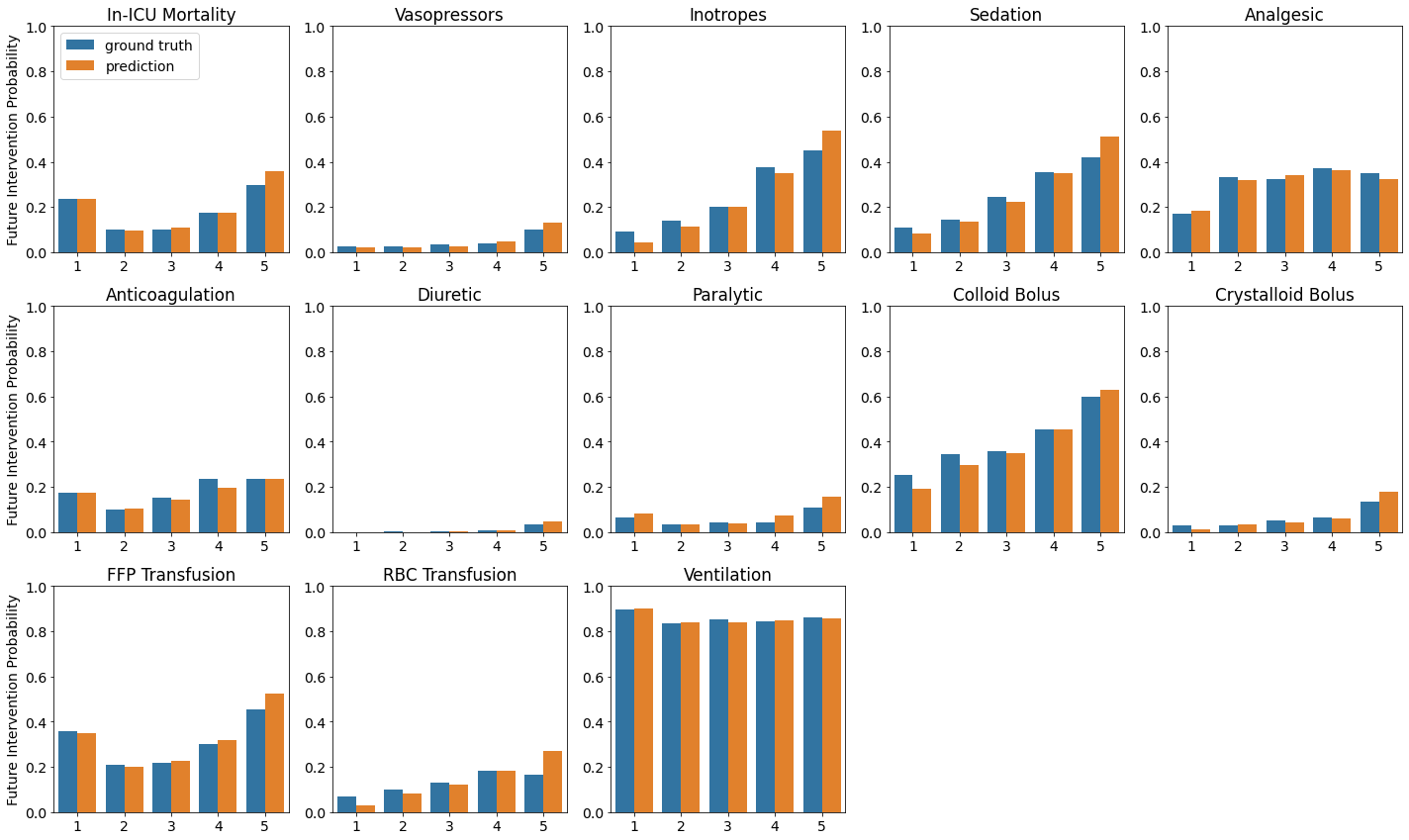}
    
    \caption{Average future intervention probability at five equally divided quantile levels (0\%-20\%, 20\%-40\%, 40\%-60\%, 60\%-80\%, 80\%-100\%) of the predicted in-ICU mortality probability.}
    
    \label{fig:8}
\end{figure}

\begin{figure}[ht]
    \centering
    \includegraphics[width=1.0\textwidth]{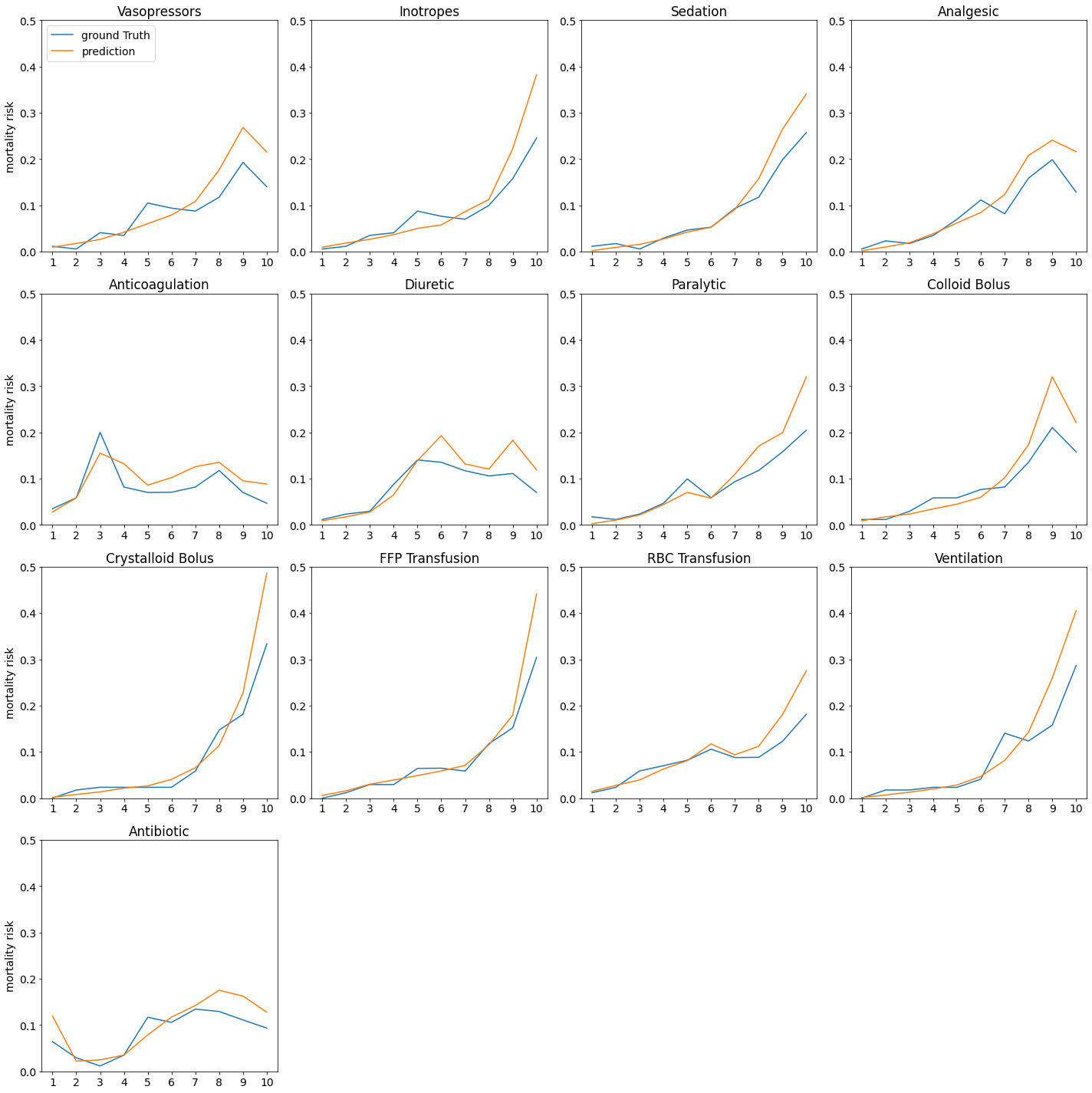}
    
    \caption{Average mortality probability at the decile of future intervention probability.}
    
    \label{fig:9}
\end{figure}

\begin{figure*}[ht]
    \centering
    \includegraphics[width=1.0\textwidth]{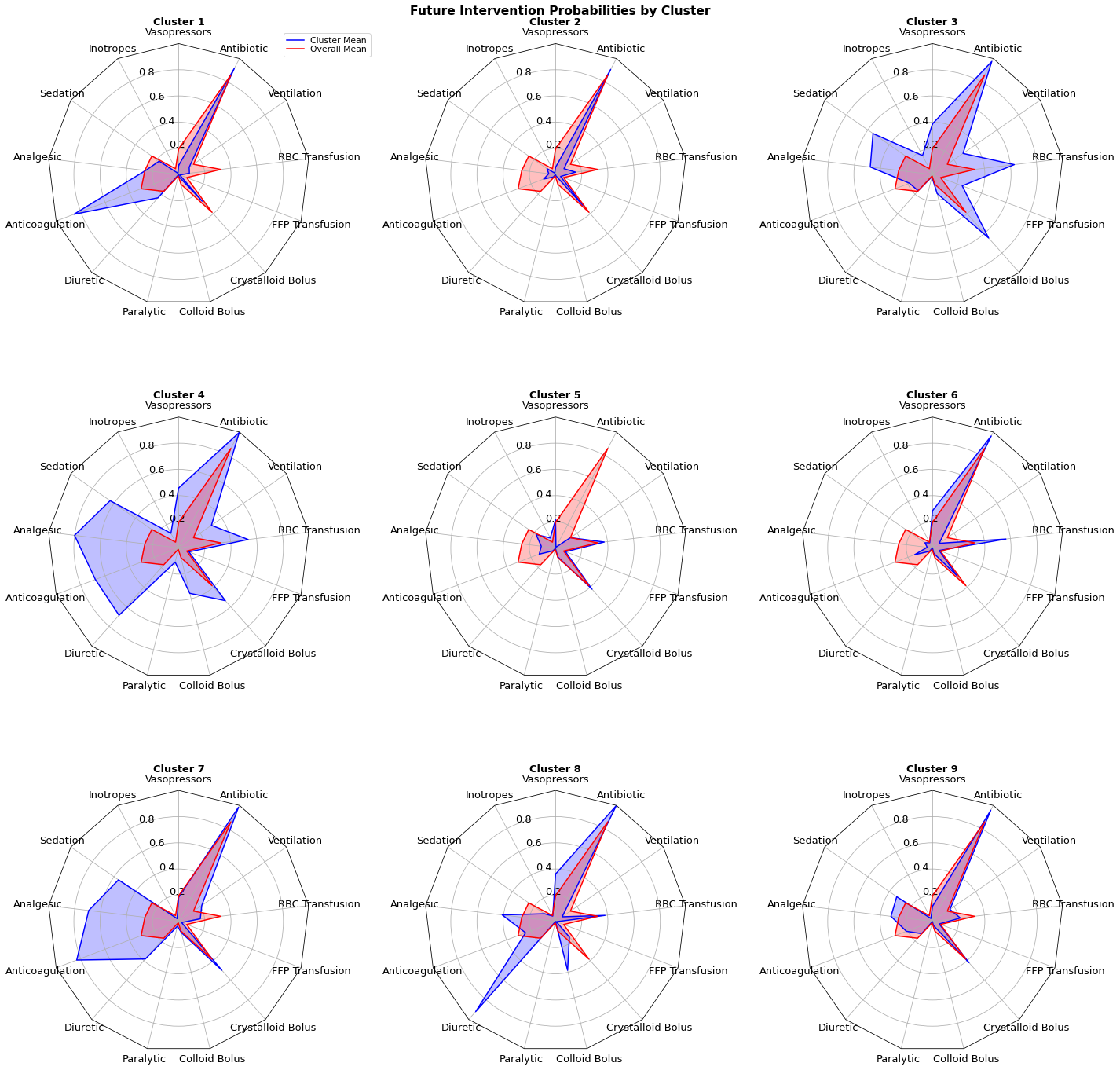}
    
    \caption{Average future intervention probabilities by cluster. Each spoke of the plot is a different intervention. Blue lines indicate mean predicted probabilities for that cluster, red lines indicate mean probabilities across all clusters.
}
    
    \label{fig:radar}
\end{figure*}

\end{document}